# Unsupervised Correlation Analysis

Yedid Hoshen
Facebook AI Research
yedid@fb.com

Lior Wolf
Facebook AI Research and Tel Aviv University
wolf@fb.com

## Abstract

*Linking between two data sources is a basic building block in numerous computer vision problems. In this paper, we set to answer a fundamental cognitive question: are prior correspondences necessary for linking between different domains?*

*One of the most popular methods for linking between domains is Canonical Correlation Analysis (CCA). All current CCA algorithms require correspondences between the views. We introduce a new method Unsupervised Correlation Analysis (UCA), which requires no prior correspondences between the two domains. The correlation maximization term in CCA is replaced by a combination of a reconstruction term (similar to autoencoders), full cycle loss, orthogonality and multiple domain confusion terms. Due to lack of supervision, the optimization leads to multiple alternative solutions with similar scores and we therefore introduce a consensus-based mechanism that is often able to recover the desired solution. Remarkably, this suffices in order to link remote domains such as text and images. We also present results on well accepted CCA benchmarks, showing that performance far exceeds other unsupervised baselines, and approaches supervised performance in some cases.*

## 1. Introduction

Computer vision seeks an understanding of the scene behind the image, mapping an image to a non-image representation. Linking visual data $X$ with an external data source $Y$ is, therefore, the defining task of computer vision. When applying machine learning tools to solve such tasks, we often consider the non-vision source $Y$ to be univariate. A more general scenario is the one in which $Y$ is also multidimensional. Examples of such view to view linking include matching a video to a concurrent audio, matching an image with its textual description, matching images from two fixed views, etc.

The classical method of tying two different domains is Canonical Correlation Analysis (CCA), which links matching vectors by maximizing the correlation between the two views. The algorithm has been generalized in many ways: regularization was added [35], kernels were introduced [2, 36, 4], versions for more than two sources were developed [46], etc. Recently, with the advent of deep learning methods, deep versions were created and showed promise [3, 52, 10, 41].

All of these methods are fully supervised and the loss terms are based on having access to matching representations in the two domains. Recently, a large body of work quickly emerged, which maps between two domains in an unsupervised way, i.e., without observing matching samples. The success of such methods is extraordinary, since there is no evidence in the existing machine learning or cognitive science literature suggesting that this would be possible.

The success, however, is limited to two visual domains [27, 57, 55, 5], or, much less commonly (and employing mostly semi-supervised learning) to two languages [53] and very recently [12, 31]. This leads to the hypothesis that the methods merely perform sophisticated style transfer, as discussed in [57], are unable to perform geometrical transformations (see discussion in [27]), focus on simple transformations [14], or compute a transformation that can be well approximated by a fixed permutation of the pixels, as demonstrated in [5].

In this work, we demonstrate the capability of matching between completely different domains in an unsupervised way. As far as we know, this is the first time that such a method is presented. By doing so we, therefore, both (i) significantly extend what is known to be possible in unsupervised learning and, (ii) provide a new practical method for many computer vision applications.

Similarly to the CCA literature, we assume the data in the two domains is already encoded, e.g., by a preexisting deep CNN for images, or a thought-vector technique for text. In order to link between the domains, our method employs multiple sub-networks of two types: to transform between domains, we employ shallow projection matrices; to link between the two data sources, we employ deeper domain confusion networks [15].

Even the computation of CCA, as is typically done with SVD or eigenvalue decomposition is ambiguous [9] since the projections is given up to a flipping of a sign. This can be easily resolved in the supervised case by making the recovered correlations positive. In the unsupervised case, correlations cannot be computed since there are no matches. In addition, there may be many more ambiguities (e.g., matching an image to a description of the same image after some transformation). In order to overcome this, we run our method $k$ times. Then, we analyze the relations between the recovered solutions by considering the first Principle Component of a data-driven similarity kernel between the solutions. The final solution emerges as the solution with the maximal value along this dimension.

Taken together, our method is able to solve a variety of computer vision tasks, inferring a missing half of an image from the given part, and link text and images.

We review related work in Sec. 2, our method UCA is detailed in Sec. 3. Experimental evaluation is presented in Sec. 4. We conclude in Sec. 5.

## 2. Related work

This paper aims to identify analogies between datasets without supervision. Analogy identification as formulated in this paper is highly related to image matching methods. As we perform matching by synthesis across domains, our method is related to unsupervised style-transfer and image-to-image mapping. In this section we give a brief overview of the most closely related works.

**Style transfer** Style transfer methods [16, 47, 26] typically receive as input a style image and a content image and create a new image that has the style of the first and the content of the second. Style is captured by local image statistics ("texture") and content is measured by the activations of a neural net classifier. The problem of image translation between domains differs since when mapping between domains, part of the content is replaced with new content that matches the target domain and not just the style.

**Generative Adversarial Networks** GAN [17] methods train a generator network $G$ that synthesizes samples from a target distribution, given noise vectors, by jointly training a second network $D$. In image mapping, the created image is based on an input image and not on random noise [27, 57, 55, 33, 45, 25], using a similar adversarial network $D$. Earlier, fully supervised, conditional GANs include generating samples from a specific class [40], based on a textual description [42, 56], or invert mid-level network activations [13].

**Unsupervised Mapping** Unsupervised mapping does not employ supervision apart from sets of sample images from the two domains. This was done very recently [27, 57, 55, 22] for image to image translation and slightly earlier for translating between natural languages [53, 31].

[45] match between the source domain and the target domain by incorporating a fixed pre-trained feature map $f$ and requiring $f$-constancy, i.e, that the activations of $f$ are the same for the input samples and for mapped samples. We do not use such assumptions in this work.

**Domain Adaptation** In this setting, we typically are given two domains, one having supervision in the form of matching labels, while the second has little or no supervision. The goal is to learn to label samples from the second domain. In [8], what is common to both domains and what is distinct is separated thus improving on existing models. In [7], a transformation is learned, on the pixel level, from one domain to another, using GANs. In [21], an unsupervised adversarial approach to semantic segmentation, which uses both global and category specific domain adaptation techniques, is proposed.

A principled way for performing domain adaptation using an adversarial network is presented by Ganin et al. [15]. The adversarial network tries to distinguish between the features extracted from samples of two the domains, after these were processed by a feature extraction network. A similar "domain confusion" network is employed in our work (although two different projects are used) and in many other recent contributions, for example, for the task of imitation learning [20].

**Canonical Correlation Analysis** (CCA) [23] is a statistical method for computing a linear projection for two views into a common space, which maximizes their correlation. CCA plays a crucial role in many computer vision applications including multiview analysis [1], multimodal human behavior analysis [44], action recognition [28], and linking text with images [30]. There are a large number of CCA variants including: regularized CCA [50], Nonparametric canonical correlation analysis (NCCA) [37], and Kernel canonical correlation analysis (KCCA) [2, 36, 4], a method for producing non-linear, non-parametric projections using the kernel trick. Recently, randomized non-linear component analysis (RCCA) [39] emerged as a low-rank approximation of KCCA. While CCA is restricted to linear projections, KCCA is restricted to a fixed kernel. Both methods do not scale well with the size of the dataset and the size of the representations. A number of methods [3, 52, 10, 41] based on Deep Learning were recently proposed that aim to overcome these drawbacks. Deep canonical correlation analysis [3] processes the pairs of inputs through two network pipelines and compares the results of each pipeline via the CCA loss.

Two contributions [54] and [51], extend DeepCCA [3] to the task of images and text matching. The first employs the same model and training process of [3] while the latter employs a different training scheme on the same architecture. Other deep CCA methods, are inspired by a family of encoding/decoding unsupervised generative models [19, 6, 34, 48, 49] that aim to capture a meaningful rep-resentation of input $x$ by applying a non-linear encoding function, decoding the encoded signal using a non-linear decoding function and minimizing the squared L2 distance between the original input and the decoded output. Needless to say, all of these methods are supervised and rely on matching samples between the domains.

## 3. Method

In this section, we describe Unsupervised Correlation Analysis (UCA). Similarly to CCA, the method projects the data from the two domains into a shared space. By using domain confusion networks, we ensure that the statistics of the two projected views are indistinguishable in the shared space. We are thus able to link between the two domains. There are several technical challenges including the domain confusion constraints and managing the instability of adversarial training.

In Sec. 3.1 we describe our architecture and training procedure. In Sec. 3.2 we describe our unsupervised network selection criterion.

### 3.1. Architecture

In order to explain our method we briefly review $CCA$. $CCA$ takes as input sets of matching views $X_i$ and $Y_i$, which are stacked as the matching columns of two matrices $X$ and $Y$. The views are assumed to be centered, i.e., $\sum_i X_i = 0$ and $\sum_i Y_i = 0$. $CCA$ sets to project both views to a common space, by employing projection matrices $W_x$ and $W_y$, such that the sum of correlations $\sum_i (W_x X_i)^\top (W_y Y_i)$ is maximized, subject to the projected data being uncorrelated, i.e., $\sum_i (W_x X_i)(W_x X_i)^\top = I$ and similarly for $W_y Y_i$.

In the unsupervised setting, we would like to learn similar projection matrices $W_x$ and $W_y$. We cannot compute the cross-domain correlations without having access to matching training samples, but we can compute the correlation of each projection. The latter allows us to apply the orthogonality constraint in order to ensure uncorrelated projections.

In addition to these projections our method also employs projections $V_x$, $V_y$ from the shared space to the two views as well as three domain confusion matrices: $D_C$ in the shared (post-projection) domain, and $D_x$, $D_y$ for the two input domains. The variables, networks, constants, and parameters that define our method are listed in Tab. 1.

We center the training data by removing the mean in each of the domains. The CCA-like projections $W_x$ and $W_y$ project the input data into $\mathbb{R}^d$:

$$C_x = W_x X \quad (1)$$

$$C_y = W_y Y \quad (2)$$

Similarly to CCA, the projections dimensions are set to be uncorrelated. In our method, this is done in a soft way with the following loss

$$L_{Orth} = \|C_x^\top C_x - I\|_2^2 + \|C_y^\top C_y - I\|_2^2 \quad (3)$$

In order to make the two projections indistinguishable, we train an adversarial network $D_c$ to distinguish between $C_x$ and $C_y$.

$$L_{D_c} = BCE(D_c(C_x), 0) + BCE(D_c(C_y), 1) \quad (4)$$

where the Binary Cross Entropy function is defined based on the sigmoid function $\sigma$ as:

$$BCE(x, z) = -z \log(\sigma(x)) - (1-z) \log(1 - \sigma(x)) \quad (5)$$

The networks $W_x$ and $W_y$, on the other hand, increase the domain confusion by minimizing the negation of $L_{D_c}$, given as:

$$L_{G_c} = BCE(D_c(C_x), 1) + BCE(D_c(C_y), 0) \quad (6)$$

The goal of the inverse projections ($V_x$, $V_y$), from the shared domain in $\mathbb{R}^d$ to the input domains is to allow domain confusion to take place in these domains, while ensuring that the inverse projection $C_x$ and $C_y$ is close to the original $X$ and $Y$:

$$L_{Rec} = \|V_x C_x, X\|_2^2 + \|V_y C_y, Y\|_2^2 \quad (7)$$

We also incorporate a "Cycle" constraint as in [57, 27, 55]. The constraint enforces a view projected to the opposite view and back to be unchanged. It provides a strong way of tying $V_x$ and $W_x$ with $V_y$ and $W_y$. The cycle constraint can be written as:

$$L_{Cyc} = \|V_x W_y V_y W_x X - X\|_2^2 + \|V_y W_x V_x W_y Y - Y\|_2^2 \quad (8)$$

The domain confusion networks $D_x$ and $D_y$ minimize the following loss:

$$L_{D_x} = BCE(D_x(V_x C_y), 0) + BCE(D_x(V_x C_x), 1) \quad (9)$$

$$L_{D_y} = BCE(D_y(V_y C_x), 0) + BCE(D_y(V_y C_y), 1) \quad (10)$$

These networks act as adversaries to $V_x$ and $V_y$, as well as to $W_x$ and $W_y$, which minimize the two losses:

$$L_{G_x} = BCE(D_x(V_x C_y), 1) \quad (11)$$

$$L_{G_y} = BCE(D_y(V_y C_x), 1) \quad (12)$$

Taken together, the "generator" networks $W_x$, $W_y$, $V_x$ and $V_y$ minimize the following loss

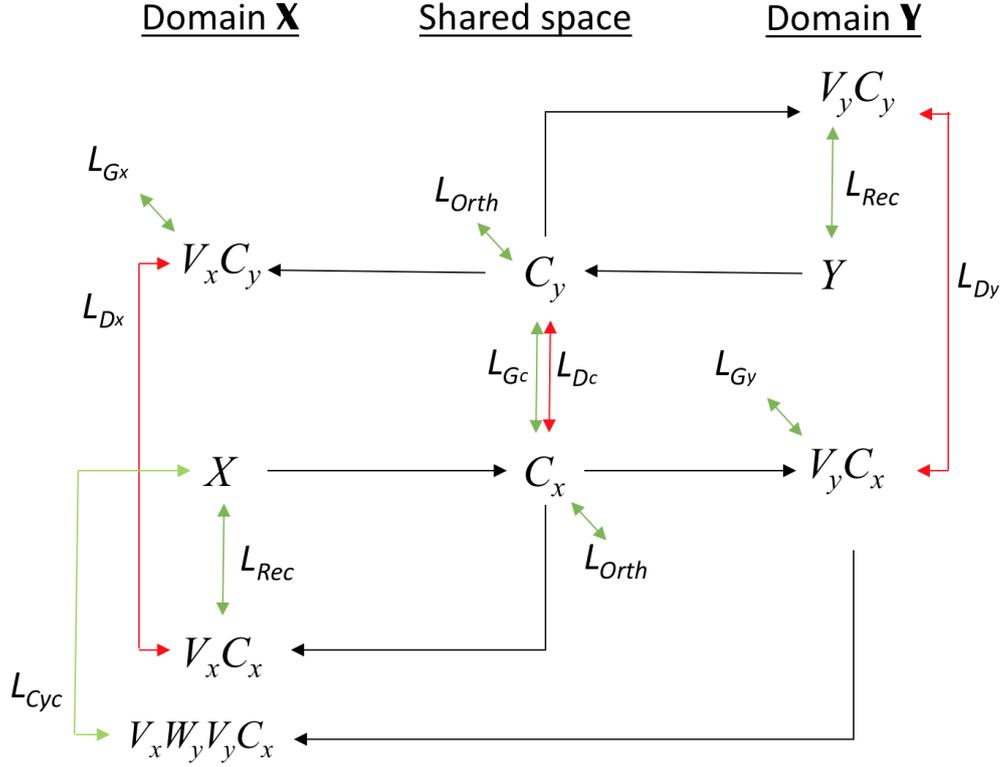

Figure 1. An illustration of the suggested method. The one-sided arrows denote projections. The double headed ones denote loss terms. Green – "generative" losses. Red – adversarial losses.

| | Symbol | Description | Computed as: |
|---|---|---|---|
| **Variables** | $X \in \mathbb{R}^{d_x \times n_x}$ | Samples in the first domain $\mathcal{X}$ | Input |
| | $Y \in \mathbb{R}^{d_y \times n_y}$ | Samples in the first domain $\mathcal{Y}$ | Input |
| | $C_X \in \mathbb{R}^{d \times n_x}$ | $X$ projected to shared space | $W_x X$ |
| | $C_Y \in \mathbb{R}^{d \times n_y}$ | $Y$ projected to shared space | $W_y y$ |
| **Networks** | $W_x : d_x \to d$ | Projection form $\mathcal{X}$ to the shared vector space | |
| | $W_y : d_y \to d$ | Projection form $\mathcal{Y}$ to the shared vector space | |
| | $V_x : d \to d_x$ | Projection form the shared space in $\mathcal{R}^d$ to $\mathcal{X}$ | |
| | $V_y : d \to d_y$ | Projection form the shared space in $\mathcal{R}^d$ to $\mathcal{Y}$ | |
| | $D_c : d \to [0,1]$ | The domain confusion discriminator $C_X$ vs. $C_Y$ | |
| | $D_x : d_x \to [0,1]$ | Discriminator for $V_x C_x$ vs. $V_x C_y$ | |
| | $D_y : d_y \to [0,1]$ | Discriminator for $V_x C_x$ vs. $V_x C_y$ | |
| **Const** | $d_x$ | Dimensionality of the input domain $\mathcal{X}$ | |
| | $d_y$ | Dimensionality of the input domain $\mathcal{Y}$ | |
| **Parameters** | $d$ | Dimensionality of the shared space | 10 |
| | $\lambda_c$ | Weight of the loss term $L_{G_c}$ | 1 |
| | $\lambda_{xy}$ | Weight of the losses $L_{G_x}, L_{G_x}, L_{D_x}, L_{D_x}$ | 1 |
| | $\lambda_{Rec}$ | Weight of the loss term $L_{Rec}$ | 1 |
| | $\lambda_{Orth}$ | Weight of the loss term $L_{Orth}$ | 1 |
| | $\lambda_{Cyc}$ | Weight of the loss term $L_{Cyc}$ | 1 |
| | $k$ | Number of reruns per experiment | 100 |

Table 1. The components of the Unsupervised Component Analysis model

$$L_G = \lambda_c L_{G_c} + \lambda_{xy} L_{G_x} + \lambda_{xy} L_{G_y} + \\ \lambda_{Rec} L_{Rec} + \lambda_{Orth} L_{Orth} + \lambda_{Cyc} L_{Cyc} \quad (13)$$

while each of the discriminator networks $D_c$, $D_x$, and $D_y$, minimizes its individual loss.

We also present two other variants, of $UCA$, $CycUCA$ and $LatentUCA$. We noticed that using all constraint together tended to decrease the total performance. $CycUCA$ is a simplified case of $UCA$, without the GAN on the latent code, and the orthogonality and reconstruction loss ($\lambda_c = \lambda_{Orth} = \lambda_{Rec} = 0$). $LatentUCA$ is the same as the full $UCA$ without the cycle constraint ($\lambda_{Cyc} = 0$).

We train all losses using mini-batch stochastic gradient descent (SGD) using the ADAM optimization algorithm [29]. As typical for adversarial loss functions we use alternating optimization, where for each mini-batch we first train the discriminator loss function and then the generator loss function. We used a learning rate of $1e-2$ and decayed it by a factor of 2 after 15 epochs. Training was performed for a total of 26 epochs. The discriminator consisted of 2 hidden layers each with 2048 nodes, followed by BatchNorm layer and ReLU activation. We used weight decay of $1e-5$ for all networks.

After training we use the projection matrices $W_x$ and $W_y$ as feature extractors from views $X$ and $Y$ respectively.

### 3.2. Unsupervised Selection Among the Runs

Performing the procedure described in Sec. 3.1 suffers both from the inherent ambiguity of unsupervised learning as well as from the well-documented instabilities in generative adversarial networks. In practice, we observe that some runs yield good performance whereas others result is projection to uncorrelated dimensions. Furthermore, since our constraints deal with the distribution level and only indirectly encourage positive correlations, we occasionally find runs that have negative correlations.

We have found that running training multiple times results in a number of runs with good performance. The challenge is being able to select the best performing runs. If we have access to an aligned (perhaps small) validation set, it can be used for choosing the best performing runs in terms of correlation or AUC. We denote this method: the Oracle. In the completely unsupervised setting, however, we do not have access to any aligned data[1]. Simple averaging was found in our experiments to be sub-optimal. Simple heuristics based on training loss have not yielded successful outcomes.

Our proposed method is based on the idea that although we do not have ground truth labels, projections that are cor-

---

[1]For testing purposes only, the performance is measured using matching points. However this cannot be done for a validation split.

---

**Algorithm 1** Unsupervised Correlation Analysis

**Require:** Unmatched samples from two domains $X$, $Y$, and the parameters listed in Tab. 1.
1: $X = \text{removeMean}(X)$
2: $Y = \text{removeMean}(Y)$
3: **for** $j \in 1 \dots k$ **do**
4:     Obtain $W_x^{(j)}$ and $W_y^{(j)}$ by minimizing Eq. 13 over $W_x, W_y, V_x$, and $V_y$, while minimizing Eq. 4, 9 10 over $D_c, D_x$ and $D_y$, respectively.
5: **end for**
6: Synthetic pairs: $(x_i, \tilde{y}_i = V_y^{(j)} W_x^{(j)} x_i) : x_i \in X$, $j$ randomized for each pair.
7: **for** $i \in 1 \dots n$ **do**
8:     **for** $j \in 1 \dots k$ **do**
9:         $M^{(j)}[i] = (W_x^{(j)} \hat{x}_i)^\top (W_y^{(j)} \tilde{y}_i)$
10:     **end for**
11: **end for**
12: $P = \text{largestSingularVector}(\text{removeMean}(M))$
13: **return** $W_x^{(j^*)}, W_y^{(j^*)}$ such that: $j^* = \arg\max\limits_{j \in 1 \dots k} P^\top M^{(j)}$

---

related with the ground truth should also be highly correlated with each other and therefore agree on more labels than failed uncorrelated projections ("crowd-wisdom"). Using randomly sampled pairs of projected points however, yields a very unbalanced set of pairs as the proportion of matching pairs is the inverse of the number of examples in the training set (this is the reason our problem is challenging in the first place). Instead, we synthesize matching pairs as follows. We pick a random example $x$ in $\mathcal{X}$, and projection matrices from a randomly selected run ($W_x$ and $V_y$). We then project point $x$ to its paired synthetic $\mathcal{Y}$ example $\tilde{y} = V_y W_x x$. We then treat $(x, \tilde{y})$ as a synthetic validation pair. we evaluate the similarity matrix of the projected synthetic pairs according to each run, $M_{ij} = W_x^j x_i \cdot W_y^j \tilde{y}_i$. for run $j$ and pair $i$.

In order to find the run that is most correlated with the other runs, we calculate the factors of variation of matrix $M$ using PCA and retain only the first principle component. The first PC approximates, using a rank-1 matrix, the kernel that measures the pair-wise similarities between runs. Therefore, the runs associated with the extremal factors in the first principle component are usually the most and least correlated runs. We show evidence for the effectiveness of this method in the experimental section. The entire method is listed in Alg. 1.

### 4. Experiments

Our experiments employ five datasets:

1. A toy problem, in which the first domain contains the MNIST images, and the second domain contains the

mirrored image.

2. A much more challenging problem in which the domain $X$ contains the upper half of the MNIST [32] images, and $Y$ contains the lower part.

3. The annotated bird dataset from [42], contains 11,788 images of birds with matching sentences. The dataset has predefined train and test splits which we follow. We encode the images using ResNet50 [18] and encode the sentences using the InferSent method of [11].

4. A similar dataset of flowers [56], with 8,189 matching pairs, encoded similarly.

5. The 8,000 images of the Flickr citeflicker 8k datasets. Each image was annotated by five sentences. We encode each image using the VGG network [43], and encode the sentences by employing fastICA [24] over the word2vec [38] representation. Namely, all word2vec representation of dictionary words were projected to 300D using fastICA. Then, the text associated with each image was taken as the average of all words in the five sentences describing it.

We evaluated multiple variants of our method and a set of baselines. The most related baseline is unsupervised generative adversarial methods. We also provide supervised CCA baselines to give an understanding of the upper bound for the performance on the evaluated datasets.

Two quantitative metrics are used for evaluation: (1) Correlation - average of the 1D dimension by dimension correlations of the code representations ($C_x$ and $C_y$), and (2) Area Under Curve (AUC) - We compute the similarity, in the latent space, between pairs of positive and negative matches (each pair has one sample from view X and one from view Y) and report the area under the ROC curve.

The unsupervised methods include the direct GAN baseline and our three variants: *CycUCA*, *LatentUCA*, *AllUCA*. An additional variant termed Oracle selects the run out of all *CycUCA*, *LatentUCA*, and *AllUCA* runs, which maximizes the correlation. It therefore serves as an upper bound for the potential utility of the selection method presented in Sec. 3.2.

**GAN:** This method trains a generative adversarial model directly between the source domain and target domain. It does not use a shared latent space, and the shared space effectively becomes the original target space. This method has a much larger number of parameters than our method, since the input space is of much larger dimensionality than the latent space. As its dimensionality is very high the correlation metric is not very meaningful for this method, and is not reported.

**UCA:** UCA is described in Sec. 3 with its three variant *CycUCA* and *LatentUCA* and *AllUCA*. *CycUCA* uses cycle constraints but no domain confusion loss on the latent space. *LatentUCA* has a domain confusion loss on the latent space but no cycle constraints. *AllUCA* contains all constraints.

We used a shared latent space of dimensionality $d = 10$. To calculate similarity between code vectors from different views we a correlation based similarity measure. This has been found to work better than the euclidean similarity. We compute 100 different training runs for each dataset, yielding 100 different projection matrices from each of the views to the shared space. For each set of projections we calculate the similarity scores on a fixed set of pairs of samples from the two views, resulting in the matrix $M$. We compute the PCA decomposition of the matrix, and represent every set of projections as the value of the top principle component. We have found that the solution (run) with the largest AUC and correlation typically has either the maximum or minimum first principle component among all solutions. We therefore report the AUC and correlation of the classifier with extremal value of the first principle components on the train set. For HalfMNIST, the second component was found to work better for all methods.

We also evaluate two supervised methods. We stress that they are only used to upper bound the performance obtainable by our method, and are not directly comparable due to inherent supervision.

**Regularized CCA (CCA):** CCA is an established technique for finding an optimal shared latent space given supervised data from two views; it maximizes correlation between the two views while enforcing orthonormality in the code space. The regularized version [50] generalizes much better, and we report results for the best regularization coefficient found.

**Supervised UCA (SupUCA):** In order to test the expressiveness of our architecture, we evaluate UCA with supervised data. Several changes are performed to accommodate the new setting. the domain confusion terms are removed. Euclidean loss terms are added for matching between the following three pairs: $(V_y W_x X_i, Y_i)$, $(V_x W_y Y_i, X_i)$, $(W_x X_i, W_y Y_i)$.

### 4.1. Results

AUC and Correlation results can be seen in Tables. 2 and 3. All UCA and supervised methods performed well on MNIST Flipped digits. This is the simplest of the tasks, as the transformation can be expressed exactly by a sparse matrix. For such simple tasks where the distributions of train and test match exactly no supervision is required to achieve perfect matching. The GAN method did not achieve a high AUC on this task, but did far better on this task than all others, which are more challenging.

MNIST Halves is a significantly harder task than MNIST Flipped. The transformation function between the top half and the bottom half or vice versa cannot perfectly reconstruct the target image. We can, however, see from the supervised baselines that a correlated latent space can be learned with near perfect accuracy given correspondences. On this task *LatentUCA* has outperformed all other unsupervised methods.

|         | Dataset |  |  |  |  |
| Method | MNIST Flipped | MNIST Halves | Birds [42] | Flowers | Flickr8k |
| --- | --- | --- | --- | --- | --- |
| *GAN* | * | * | * | * | * |
| *CycUCA* | 0.96/0.95 | 0.13/0.11 | 0.36/0.33 | 0.06/0.07 | 0.38/0.38 |
| *LatentUCA* | 1.00/1.00 | 0.25/0.25 | 0.38/0.32 | 0.10/0.10 | 0.23/0.23 |
| *AllUCA* | 1.00/1.00 | 0.23/0.23 | 0.28/0.26 | 0.05/0.02 | 0.00/0.00 |
| *Oracle* | 1.00/1.00 | 0.34/0.34 | 0.57/0.56 | 0.26/0.23 | 0.46/0.47 |
| *CCA10* | 1.00/0.99 | 0.91/0.88 | 0.88/0.70 | 0.96/0.68 | 0.79/0.73 |
| *CCA50* | 1.00/1.00 | 0.55/0.52 | 0.69/0.38 | 0.77/0.37 | 0.64/0.51 |
| *SupUCA* | 0.90/0.90 | 0.90/0.90 | 0.83/0.74 | 0.90/0.74 | 0.80/0.75 |

Table 2. Mean Cross-Domain Correlation (% train/test) on the [42] dataset. *CCA10* and *CCA50* differ in the dimensionality of the latent space. The *UCA* methods all use a latent space of size 10. * The correlation, in the sense of the other methods, is ill defined for *GAN* since there is no latent space.

|         | Dataset |  |  |  |  |
| Method | MNIST Flipped | MNIST Halves | Birds [42] | Flowers | Flickr8k |
| --- | --- | --- | --- | --- | --- |
| *GAN* | 0.69/0.69 | 0.55/0.55 | 0.54/0.54 | 0.54/0.53 | 0.58/0.58 |
| *CycUCA* | 1.00/1.00 | 0.60/0.59 | 0.72/0.70 | 0.52/0.55 | 0.76/0.76 |
| *LatentUCA* | 1.00/1.00 | 0.73/0.73 | 0.80/0.76 | 0.60/0.59 | 0.68/0.68 |
| *AllUCA* | 1.00/1.00 | 0.67/0.67 | 0.74/0.74 | 0.57/0.53 | 0.51/0.50 |
| *Oracle* | 1.00/1.00 | 0.74/0.74 | 0.86/0.83 | 0.68/0.70 | 0.82/0.81 |
| *CCA10* | 1.00/1.00 | 0.96/0.96 | 0.94/0.92 | 0.94/0.94 | 0.93/0.91 |
| *CCA50* | 1.00/1.00 | 0.98/0.98 | 0.97/0.95 | 0.97/0.96 | 0.97/0.97 |
| *SupUCA* | 0.98/0.98 | 0.95/0.95 | 0.98/0.94 | 0.99/0.95 | 0.97/0.95 |

Table 3. Cross-Domain Matching AUC (% train/test).

One may question whether *CycleGAN*/*DiscoGAN* architectures may simply solve this task, due to their careful design for unsupervised image mapping (and consequently matching). We therefore evaluate *DiscoGAN* on the MNIST Halves task, which is an image-to-image task and is challenging in the unsupervised setting (as opposed to the Flipped task). We use *DiscoGAN* rather than *CycleGAN*, since it uses an encoder-decoder rather than U-NET architecture, and is thus designed for global rather than spatially local transformations. Some example results are depicted in Fig. 2. As can be seen from the images, the *DiscoGAN* method creates realistic looking half-digits. However, the *DiscoGAN* results are not similar to the ground truth and do not perform well on this task, providing evidence for its difficulty.

Our results for MNIST-Halves are shown in Fig. 3. We show the top half (domain $X$), the matching bottom half

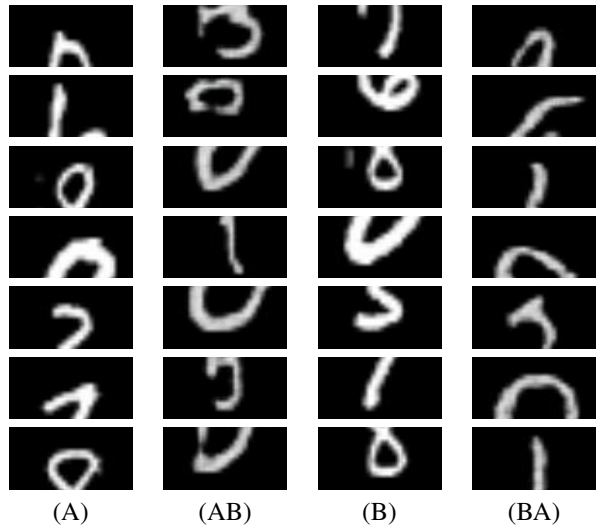

(A) (AB) (B) (BA)

Figure 2. Each row depicts the DiscoGAN mapping (AB) of the top half of an MNIST [32] digit (A) to its matching bottom half (B). Mapping (BA) is also shown between the bottom half (B) to the top half (A).

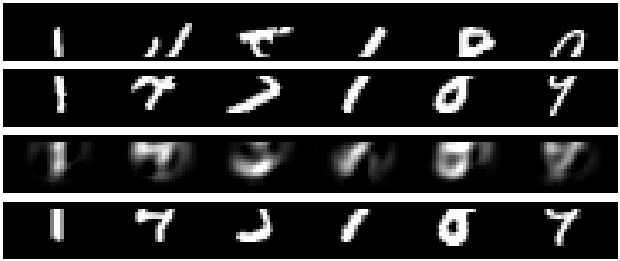

Figure 3. Shown, row by row, are the input sample [32] (top half of digit), the ground truth target (bottom half), the estimated target, and the retrieved target. The blurriness of the estimated sample is due to the 10D bottleneck but both the estimation and the top-retrieval match the target well.

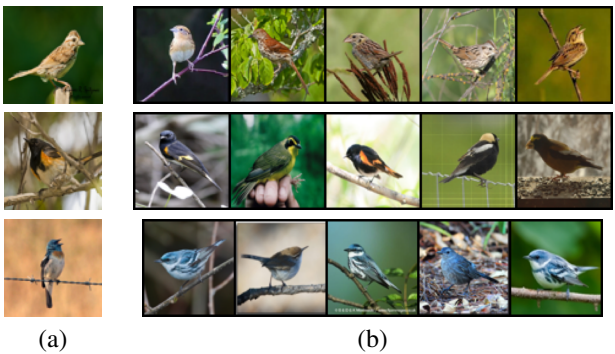

(a)                            (b)

Figure 4. (a) Ground truth image and (b) top-5 retrievals for the matching descriptions on the Birds [42] dataset which are, row by row: "small whit and brown spotted bird with a small beak ", "this bird has wings that are black and orange and has a black throat", "a round bird with a blue head, orange throat cream belly, and black wings"

(domain $Y$), the recovered image in domain $Y$, and the retrieval in domain $Y$. Note that in this experiment, it is possible to show the recovered image since the raw image pixels are used as the features. The recovered image is blurred since the reconstruction goes through a bottleneck of size 10. However, using it for retrieving from the target domain results in results that are both correct and sharp.

All $UCA$ variants performed well on the birds dataset, with some advantage to $LatentUCA$. $GAN$ has not performed well on this dataset. We hypothesize that this is due to the size of the projection matrix and the lack of low dimensional shared space in the GAN baseline method. Sample retrieval results, using $LatentUCA$ for the text to birds benchmark are shown in Fig. 4.

Flowers proved quite challenging for all unsupervised methods. *LatentUCA* was the best performer on this dataset as well, but its performance was still quite low. *CycGAN* did particularly poorly, this might be because the utility of the cycle constraint is much reduced when the reconstruc-tion quality is low. Sample retrieval results for *LatentUCA* on this benchmark are shown in Fig. 5.

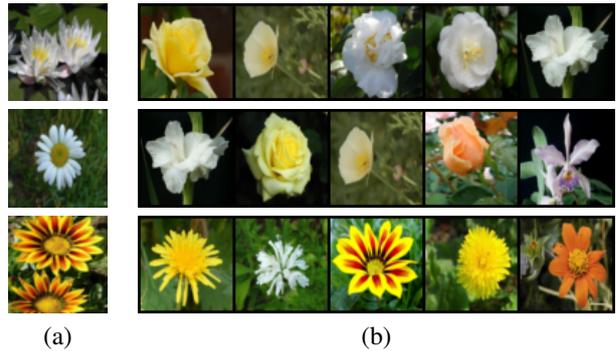

(a)                            (b)

Figure 5. (a) Ground truth image and (b) top-5 retrievals for the matching descriptions, on Flowers [56], which are, row by row: "petals are white in color, many stamne with anthers that are yellow in color", "the petals are very uniform, white, and oblong, forming a circle", "the petals of this flower are yellow with a short stigma."

*CycUCA* and *LatentUCA* performed well on Flickr8k, whereas *AllUCA* failed. We actually found that *AllUCA* had several high accuracy runs, but the unsupervised selection step did not find them.

In all experiments, the encoder-decoder architecture ensures that we are able to project views into a space where related data are highly correlated. This is not possible in the input space as done by the *GAN* method, due to high dimensionality.

Consensus-based methods were effective for unsupervised run selection, when combined with our PCA selection method. The difference in performance in comparison the to Oracle Method in most cases, is only moderate.

## 5. Conclusions

In this paper, we presented a method for linking unpaired samples from two sources. This contributes towards answering an important cognitive question, is it necessary to have prior correspondences to learn the links between domains? We presented a method to solve this important task and have shown our method to work well on image-to-image and image-to-text unsupervised linking. Linking image to text in an unsupervised way is an unintuitive result and has not been successfully performed before.

Learning unsupervised CCA is challenging and multiple constraints were found to be important for its success (orthogonality, cycle, autoencoder and domain confusion). Another challenge we overcame was training instability, which was tackled by training multiple runs and selecting the top solution using a consensus-based method.